\documentclass[journal]{IEEEtran}
\usepackage{amsmath,amsfonts,bm}

\newcommand{\abp}{{\textit{approximate best-preferred agent}}}
\newcommand{\br}{{\textit{oracle}}}
\newcommand{\abr}{{\textit{approx\_oracle}}}

\def\eqref#1{equation~\ref{#1}}
\def\1{\bm{1}}

\def\vw{{\bm{w}}}

\DeclareMathAlphabet{\mathsfit}{\encodingdefault}{\sfdefault}{m}{sl}
\SetMathAlphabet{\mathsfit}{bold}{\encodingdefault}{\sfdefault}{bx}{n}

\def\gA{{\mathcal{A}}}

\def\gE{{\mathcal{E}}}

\def\gG{{\mathcal{G}}}

\def\gJ{{\mathcal{J}}}

\def\gM{{\mathcal{M}}}
\def\gN{{\mathcal{N}}}
\def\gO{{\mathcal{O}}}
\def\gP{{\mathcal{P}}}

\def\gS{{\mathcal{S}}}

\def\gU{{\mathcal{U}}}
\def\gV{{\mathcal{V}}}

\newcommand{\E}{\mathbb{E}}

\usepackage{url}
\usepackage{algorithm}
\usepackage{graphicx}
\usepackage{subcaption}
\usepackage{algpseudocode} % Use algpseudocode package for algorithmic environment
\usepackage{amsmath} % Recommended for mathematical symbols
\usepackage{float}
\usepackage{array,booktabs,makecell,multirow}
\usepackage{xspace}
\usepackage{thmtools, thm-restate}
\newtheorem{theorem}{Theorem}[section]

\newtheorem{definition}[theorem]{Definition}

\newcommand{\algo}{HOLA-Drone\xspace} 

\newcommand{\algoR}{$\text{HOLA-Drone}_R$\xspace}

\newcommand{\bt}[1]{\textit{\textbf{#1}}}

\ifCLASSINFOpdf
\else
\fi
\begin{document}
%
% paper title
% Titles are generally capitalized except for words such as a, an, and, as,
% at, but, by, for, in, nor, of, on, or, the, to and up, which are usually
% not capitalized unless they are the first or last word of the title.
% Linebreaks \\ can be used within to get better formatting as desired.
% Do not put math or special symbols in the title.
\title{HOLA-Drone: Hypergraphic Open-ended Learning for Zero-Shot Multi-Drone Cooperative Pursuit}
%
%
% author names and IEEE memberships
% note positions of commas and nonbreaking spaces ( ~ ) LaTeX will not break
% a structure at a ~ so this keeps an author's name from being broken across
% two lines.
% use \thanks{} to gain access to the first footnote area
% a separate \thanks must be used for each paragraph as LaTeX2e's \thanks
% was not built to handle multiple paragraphs
%Dengyu Zhang, Junfan Chen, Ying Wen, Qingrui Zhang, Wei Pan ~\IEEEmembership{Student Member,~IEEE,}

\author{Yang~Li,
        Dengyu~Zhang,
        Junfan~Chen,
        Ying~Wen,  Qingrui~Zhang, Shaoshuai Mou, Wei~Pan% <-this % stops a space
\thanks{Yang Li, Junfan Chen, and Wei Pan are with the Department of Computer Science, The University of Manchester, Manchester, UK.}
\thanks{Dengyu Zhang and Qingrui Zhang are with the School of Aeronautics and Astronautics, Sun Yat-sen University, Shenzhen, China.}
\thanks{Ying Wen is with the School of Electronic, Information and Electrical Engineering, Shanghai Jiao Tong University, Shanghai, China.}% <-this % stops a space
\thanks{Shaoshuai Mou is with School of Aeronautics and Astronautics, Purdue University, USA.}
}

% note the % following the last \IEEEmembership and also \thanks - 
% these prevent an unwanted space from occurring between the last author name
% and the end of the author line. i.e., if you had this:
% 
% \author{....lastname \thanks{...} \thanks{...} }
%                     ^------------^------------^----Do not want these spaces!
%
% a space would be appended to the last name and could cause every name on that
% line to be shifted left slightly. This is one of those "LaTeX things". For
% instance, "\textbf{A} \textbf{B}" will typeset as "A B" not "AB". To get
% "AB" then you have to do: "\textbf{A}\textbf{B}"
% \thanks is no different in this regard, so shield the last } of each \thanks
% that ends a line with a % and do not let a space in before the next \thanks.
% Spaces after \IEEEmembership other than the last one are OK (and needed) as
% you are supposed to have spaces between the names. For what it is worth,
% this is a minor point as most people would not even notice if the said evil
% space somehow managed to creep in.

% The paper headers
\markboth{Journal of \LaTeX\ Class Files,~Vol.~14, No.~8, August~2015}%
{Shell \MakeLowercase{\textit{et al.}}: Bare Demo of IEEEtran.cls for IEEE Journals}
% The only time the second header will appear is for the odd numbered pages
% after the title page when using the twoside option.
% 
% *** Note that you probably will NOT want to include the author's ***
% *** name in the headers of peer review papers.                   ***
% You can use \ifCLASSOPTIONpeerreview for conditional compilation here if
% you desire.

% If you want to put a publisher's ID mark on the page you can do it like
% this:
%\IEEEpubid{0000--0000/00\$00.00~\copyright~2015 IEEE}
% Remember, if you use this you must call \IEEEpubidadjcol in the second
% column for its text to clear the IEEEpubid mark.

% use for special paper notices
%\IEEEspecialpapernotice{(Invited Paper)}

% make the title area
\maketitle

% As a general rule, do not put math, special symbols or citations
% in the abstract or keywords.
\begin{abstract}
Zero-shot coordination (ZSC) is a significant challenge in multi-agent collaboration, aiming to develop agents that can coordinate with unseen partners they have not encountered before. Recent cutting-edge ZSC methods have primarily focused on two-player video games such as OverCooked!2 and Hanabi. In this paper, we extend the scope of ZSC research to the multi-drone cooperative pursuit scenario, exploring how to construct a drone agent capable of coordinating with multiple unseen partners to capture multiple evaders. We propose a novel Hypergraphic Open-ended Learning Algorithm (HOLA-Drone) that continuously adapts the learning objective based on our hypergraphic-form game modeling, aiming to improve cooperative abilities with multiple unknown drone teammates. To empirically verify the effectiveness of HOLA-Drone, we build two different unseen drone teammate pools to evaluate their performance in coordination with various unseen partners. The experimental results demonstrate that HOLA-Drone outperforms the baseline methods in coordination with unseen drone teammates. Furthermore, real-world experiments validate the feasibility of HOLA-Drone in physical systems. Videos can be found on the project homepage~\url{https://sites.google.com/view/hola-drone}.

\end{abstract}

% Note that keywords are not normally used for peerreview papers.

% For peer review papers, you can put extra information on the cover
% page as needed:
% \ifCLASSOPTIONpeerreview
% \begin{center} \bfseries EDICS Category: 3-BBND \end{center}
% \fi
%
% For peerreview papers, this IEEEtran command inserts a page break and
% creates the second title. It will be ignored for other modes.
\IEEEpeerreviewmaketitle

\section{Introduction}

In multirobot system, the problem of cooperative drone pursuit plays a crucial role in various applications such as surveillance and urban security~\cite{chung2011search,ZhangDACOOP2023,queralta2020collaborative}. 
A key challenge in these settings arises when a drone must coordinate with previously unseen teammates. 
This is where zero-shot coordination (ZSC) comes into play, which allows for efficient collaboration with new partners and has gained considerable attention in cooperative AI~\cite{otherplay,carroll2019utility,FCP,TrajD,MEP,COLE,yuan2023survey}. 
In this paper, we address the \bt{zero-shot cooperative multi-drone pursuit problem}, where a learner drone is placed in a cooperative scenario with multiple unseen drone teammates to pursue and capture multiple evaders. The learner drone must rapidly coordinate with its teammates without modifying its fixed policy from the training phase~\cite{otherplay}.

% Efficient coordination with previously unseen partners, a problem also known as zero-shot coordination (ZSC), remains a central challenge in cooperative AI~\cite{otherplay,carroll2019utility}. 
% The most current works in ZSC~\cite{otherplay,carroll2019utility,FCP,TrajD,MEP,COLE} focus on two-player video games such as Hanabi~\cite{bard2020hanabi} and Overcooked!2~\cite{carroll2019utility}. 
% However, ZSC is not limited to video games; it is a critical challenge in real-world robotic applications such as autonomous drone swarms~\cite{chen2020toward}, search and rescue missions~\cite{queralta2020collaborative}, and collaborative warehouse robots~\cite{bolu2021adaptive}. 
% Multi-robot cooperation in the real world exacerbates the challenge of efficient coordination among unseen partners due to the need to manage multiple agents, navigate a large action space, account for hardware variability, and contend with unpredictable environments.

% In particular, the cooperative drone pursuit problem is a critical issue in multi-robot cooperation, playing a crucial role in various applications such as surveillance and city security~\cite{chung2011search,ZhangDACOOP2023}.
% In this paper, we focus on the \bt{zero-shot cooperative multi-drone pursuit problem}, where a drone agent, termed the learner, is placed in a cooperative scenario with multiple unseen drone teammates to capture multiple evaders. The learner must quickly coordinate with its teammates without updating its policy, which remains fixed from the training phase~\cite{otherplay}.

Recent advances in multi-robot pursuit research can be categorized into three main approaches: rule-based heuristic methods, differential game theoretical methods, and learning-based methods. Rule-based heuristic methods are primarily inspired by biological behaviors, aiming to imitate hunting and pursuit actions observed in nature~\cite{madden_modeling_2011, muro_wolf-pack_2011, angelani_collective_2012, zhou_cooperative_2016, pierson_intercepting_2017, shah_multi-agent_2019}. However, these methods face limitations when dealing with complex tasks, such as those involving evader movement advantages and complex environments. The second category of methods utilizes differential game theory to address the multi-robot pursuit problem by deriving theoretically optimal strategies. These methods, however, often require precise state transition equations, leading to reduced performance under conditions of uncertainty. Lastly, learning-based methods have significant advantages in enhancing distributed cooperative pursuit abilities across various environments and tasks~\cite{matignon_hysteretic_2007, li_robust_2019, qi_cascaded_2024, de_souza_decentralized_2021}. Nevertheless, even the most advanced methods, including learning-based approaches, struggle to handle scenarios involving collaboration with unseen partners in multi-robot pursuit task.

The most current methods for solving the zero-shot coordination problem focus on two-player video games~\cite{bard2020hanabi,carroll2019utility}. 
Conventional self-play and its subsequent methods~\cite{SP, carroll2019utility, otherplay} aim to improve cooperative ability by playing with a copy or permutation of itself.
Cutting-edge approaches often involve pre-training diverse strategy populations and then training a common best-response to these pretrained diverse agents to enhance coordination with various agents~\cite{FCP, TrajDi, Canaan2022Hanabi, MEP}. 
Furthermore, recent research~\cite{charakorn2023generating, COLE} has focused on improving compatibility within the population to further enhance cooperative abilities.
However, these methods are typically designed for two-player video games in 2D simulators, rather than for multi-agent collaboration in the physical world.

In this paper, we propose a novel \bt{H}ypergraphic \bt{O}pen-ended \bt{L}earning \bt{A}lgorithm (\bt{\algo}) to address the multi-drone zero-shot coordination problem in cooperative pursuit tasks.
To the best of our knowledge, this is the first work to formulate the cooperative multi-drone pursuit task as a ZSC problem within the framework of a decentralized partially observable Markov decision process (Dec-POMDP). 
Most existing works on Dec-POMDPs in multi-drone pursuit~\cite{MultiZhang2022,ZhangDACOOP2023,10417793} have primarily focused on coordination among agents with known teammates. By introducing the ZSC setting, we open the door to more adaptable and robust solutions, allowing drone teams to effectively collaborate with unseen partners in a variety of unknown or evolving environments.
We further propose the novel Hypergraphic-Form game and corresponding concepts to capture the cooperative interaction relationships among multiple agents in a hypergraph, allowing us to effectively evaluate the cooperative abilities of each agent. 
These concepts are incorporated into the open-ended learning framework~\cite{Srivastava2012Comtinually,Team2021OpenEndedLL,Meier2022Open}, which automatically adapts learning objectives to continuously improve the learner’s cooperative ability with various teamamtes, unlike most current methods that rely on fixed predefined objective functions.
To evaluate the ZSC ability with multiple unseen drone agents, we first construct two sets of drone agents with different algorithms and varying levels of cooperative ability as unseen partners.
We then conduct a series of experiments to verify the effectiveness of \algo compared to baselines when coordinating with unseen drone partners sampled from these pre-built pools. Additionally, we directly deploy the learned policies in real-world Crazyflie drones to justify the feasibility of \algo in physical systems.

Our contributions can be summarized as threefold:
\bt{(1)} To the best of our knowledge, we are the first to frame the cooperative pursuit task as a zero-shot coordination problem, enabling drone to collaborate effectively with unseen partners in evolving environments. 
\bt{(2)} We propose a hypergraphic open-ended learning algorithm to continuously improve the learner's cooperative ability with multiple teammates in the complex multi-drone pursuit task, moving beyond the two-player video games typically used in current ZSC research.
\bt{(3)} The simulation and real-world experiments, conducted with a set of previously unseen drone partners, demonstrate the effectiveness and feasibility of the proposed algorithm \algo in physical world.

\begin{figure*}
    \centering
    % \vspace{-0.6cm}
    \includegraphics[width=\textwidth]{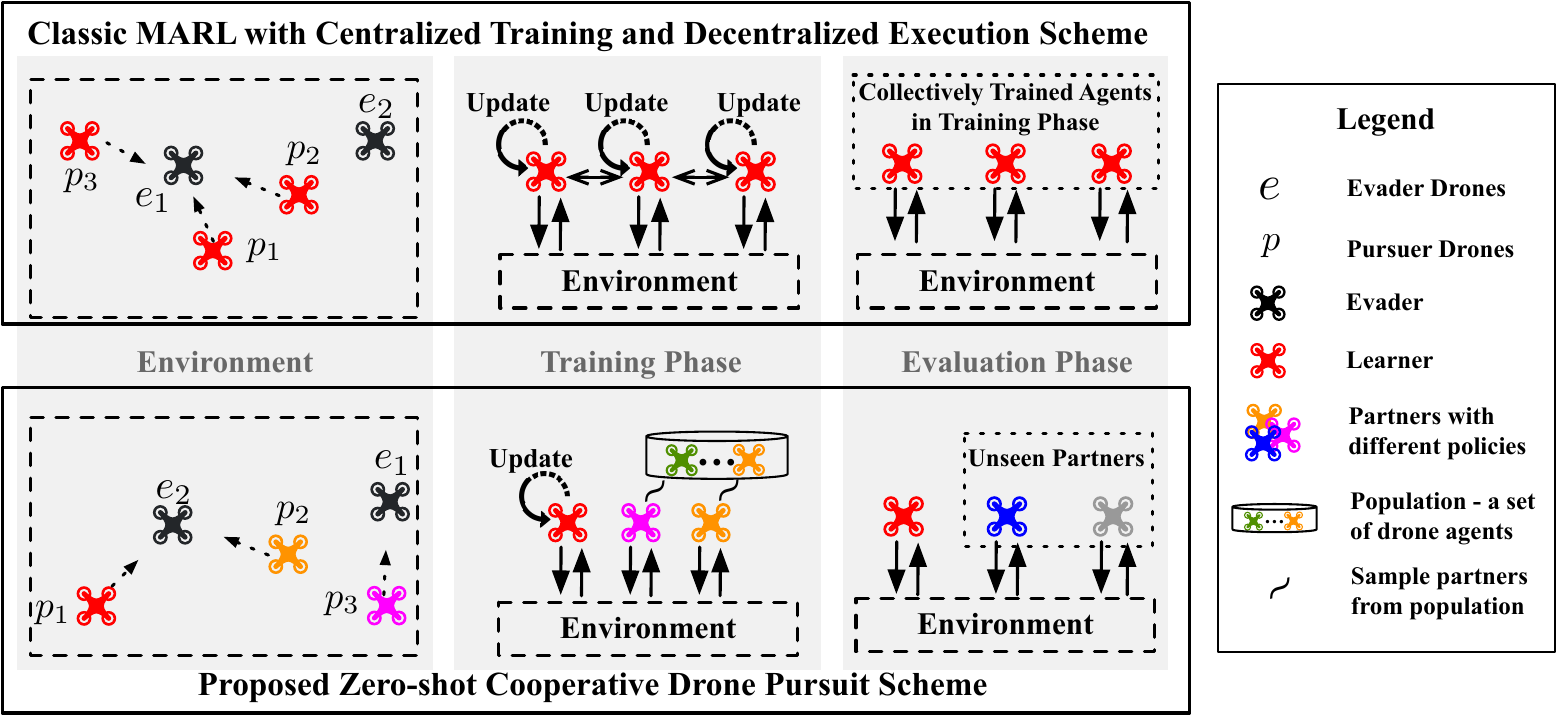}
    \caption{
\textbf{Top row: Classic multi-agent reinforcement learning (MARL) with centralized training and decentralized execution (CTDE) framework. }
- \textit{Training Phase:} All agents are updated collectively.
- \textit{Evaluation Phase:} The same agents that were involved in the training phase are deployed. The evaluation assesses the collective performance of these pre-trained agents in achieving the task using the strategies learned during training.
\textbf{Bottom row: 
Proposed zero-shot cooperative multi-drone pursuit scheme.}
- \textit{Training Phase:} A single learner agent is trained by co-playing with a set of non-learnable partners.
- \textit{Evaluation Phase:} The learner agent is required to coordinate with previously unseen partner agents that were not part of its training. 
The goal is to assess the learner's zero-shot coordination ability with any new, unseen partner without additional updating.
    }
    \label{fig:explain_zsc}
    % \vspace{-0.5cm}
\end{figure*}

\section{Related Work}
\label{app:related}
\paragraph{Multi-robot pursuit.}
One of the classic methods for solving the multi-robot pursuit problem is rule-based heuristic methods.
Heuristic rules inspired by biological behavior have been proposed to imitate hunting and pursuit actions observed in nature. Most heuristic rules are implemented using artificial attractive and repulsive forces~\cite{madden_multi-robot_2010, muro_wolf-pack_2011, angelani_collective_2012}. 
For instance, \cite{janosov_group_2017} designed predictive attraction forces, obstacle repulsive forces, and teammate repulsive forces for encirclement. 
Similarly, \cite{pierson_intercepting_2017} proposed a method based on Voronoi tessellation, which is suitable for ground vehicles. However, these rule-based methods are manually designed based on the observations or experiences of the designer, which limits their applicability.
Differential game theory is another approach to solving the multi-robot pursuit problem~\cite{mu_survey_2023, garcia_geometric_2017, kothari_cooperative_2017, hayoun_two--one_2017}. 
\cite{hayoun_two--one_2017} modeled a two-on-one pursuit problem as a zero-sum game and obtained an analytical solution. \cite{kothari_cooperative_2017} considered pursuers and evaders as nonholonomic constraint systems and introduced model predictive control to minimize the evader's safe zone. 
These methods, similar to optimal control, derive strategies by maximizing the utility function based on the Hamilton-Jacobi-Bellman equation of the system. However, they require precise state transition equations, resulting in reduced performance under uncertainty and unknown environments.
Learning-based methods have been proposed to enhance distributed cooperative pursuit abilities in various environments and tasks~\cite{matignon_hysteretic_2007, li_robust_2019, qi_cascaded_2024, de_souza_decentralized_2021}. \cite{de_souza_decentralized_2021} used curriculum learning and parameter sharing techniques to train collaborative intelligent agents, which initially move towards the evader, then slow down at an appropriate distance, and disperse to surround the evader. Although the learned strategy, under a well-designed reward function and substantial data, exhibits intelligence similar to biological behavior, this method lacks scalability. \cite{zhang_multi-agent_2022} proposed an attention interface to enhance interaction between agents and the environment, demonstrating better performance in a 100-on-100 collaborative pursuit task. \cite{ZhangDACOOP2023} introduced a hybrid design that integrates rule-based strategies into reinforcement learning for multi-robot pursuit, improving data efficiency and generalization.
However, most of these methods may fail to coordinate with unseen partners. To the best of our knowledge, we are the first to propose the zero-shot multi-drone cooperative pursuit problem and introduce a novel \algo algorithm to handle the problem.

\paragraph{Zero-shot coordination.} \cite{carroll2019utility} introduced a novel two-player, fully cooperative environment inspired by the popular game Overcooked, which demands challenging coordination. Their method involves learning a simple model that emulates human play, serving as a standard unseen evaluation agent within the Overcooked environment. The self-play (SP) method~\cite{SP} was implemented in this Overcooked setting to coordinate zero-shot with the unseen human proxy model~\cite{carroll2019utility}. However, SP tends to become stuck in the conventions formed between trained players, making it in a unable to cooperate with other unseen strategies~\cite{Adam2018Learning,otherplay}.
To address this issue, other-play~\cite{otherplay} was proposed, introducing permutations to one of the strategies to break the conventions formed by the self-play method~\cite{SP, carroll2019utility}. However, this approach may revert to self-play if the game or environment lacks symmetries or has unknown symmetries.
Recent ZSC research is mainly inspired by population-based training (PBT), which improves adaptability by fostering cooperation with multiple strategies within a population~\cite{carroll2019utility}. However, PBT does not explicitly maintain diversity, thus failing to coordinate with unseen partners~\cite{FCP}. To address this limitation and achieve the goal of ZSC, cutting-edge methods emphasize pre-training diverse strategy populations~\cite{FCP,TrajDi} or applying handcrafted techniques~\cite{Canaan2022Hanabi,MEP} to excel in cooperative games by optimizing objectives within these populations.
Furthermore, mechanisms such as coevolution and combinatorial generalization have been introduced to enhance generalization ability~\cite{Xue2022Heter,Mahajan2022Gen}. Recent research~\cite{charakorn2023generating,COLE} has also focused on improving compatibility within the population to enhance cooperative abilities. 
However, most of these methods address the ZSC problem in two-player video games such as Hanabi~\cite{bard2020hanabi} and Overcooked!2~\cite{carroll2019utility}. In this work, we extend the scope to multiple agents beyond two players, tackling more complex real-world cooperative multi-drone pursuit problems instead of video games in simulator.

\section{Problem Formulation}
In this paper, we focus on the multi-drone cooperative pursuit task, where pursuit drones aim to capture faster evaders, within a confined environment containing obstacles. 
An evader $e$ is considered captured by a drone $p$ if the distance $d_{p, e}$ between $p$ and $e$ is less than a predefined capture threshold $d_c$. 
Furthermore, a collision is defined to occur if the distance between two drones is less than $\kappa$ or if the distance between a drone and an obstacle is less than $d_s$, where $d_s$ is the safe radius for drones. Thus, the objective of the pursuers is to capture all evaders without collisions.

The zero-shot coordination problem arises in scenarios where there is no opportunity to update the fixed policy established during training when coordinating with unseen partners~\cite{otherplay}. Fig.~\ref{fig:explain_zsc} compares two frameworks: the standard multi-agent reinforcement learning (MARL) with centralized training and decentralized execution (CTDE) and the zero-shot cooperative multi-drone pursuit scheme.
In the upper row of Fig.~\ref{fig:explain_zsc}, the training and evaluation phases of MARL with CTDE are illustrated along with the corresponding environment schematic diagram. In the CTDE scheme, all controlled agents are updated collectively during the training phase using a shared policy or a centralized critic with access to centralized information. During the evaluation, the same agents trained in the training phase are deployed to assess their collective performance in achieving the task using the learned strategies.
However, the zero-shot cooperative multi-drone pursuit framework, shown in the bottom row of Fig.~\ref{fig:explain_zsc}, is designed to train a single learner agent to co-play with other non-learnable partners. The evaluation phase for this scheme is notably different: the learner agent is required to coordinate with unseen partners, unlike in the classic MARL scheme where partners are collectively trained together. This approach assesses the learner’s zero-shot coordination ability—its ability to work effectively with any new, unseen partner without additional updates.

\begin{figure*}[t]
    \centering
    \includegraphics[width=\textwidth]{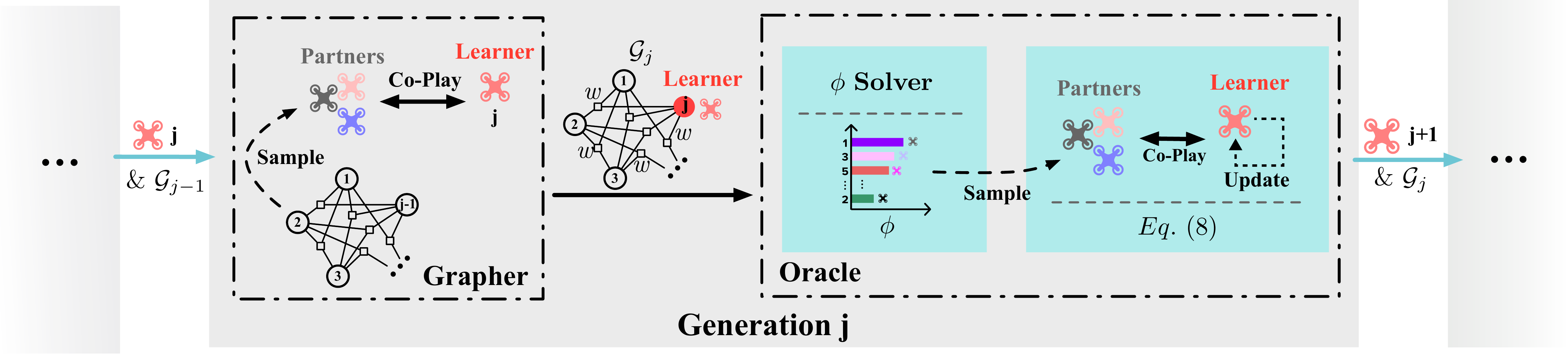}
    \caption{Hypergraphic Open-ended Learning Algorithm: Detailed illustration of a single generation within the open-ended learning phase, including the Grapher and Oracle modules.}
    \label{fig:model}
    % \vspace{-5mm}
\end{figure*}

The zero-shot multi-drone pursuit problem can be effectively modeled as a decentralized partially observable Markov decision process (Dec-POMDP). 
A Dec-POMDP, denoted as $\gM$, is defined by the tuple $(\gS, \gN, \gA, \gP, r, \gO, \gamma, T)$. Here, $\gN$ represents the set of drone agents, which includes both the pursuers ($\gN_p$) and the evaders ($\gN_e$). Specifically, $\gN_p$ consists of the learner ($\gN_1$) and its co-players ($\gN_{-1}$). Additionally, $\gS$ denotes the joint-state space, while $\gA = \times_{j=1}^k A^j$ represents the joint-action space, where $k$ is the team size. $\gP$ and $\gO$ are the transition and observation functions, respectively. The reward function is denoted by $r$, $\gamma$ is the reward discount factor, and $T$ represents the task horizon.
In the zero-shot multi-drone pursuit task, the policies of the learner’s teammates and the evaders are often pre-trained or pre-defined. At the beginning of each episode, the learner’s teammates $\gN_{-1}$ are sampled from a population $\gU$. At time $t > 0$, the Dec-POMDP is in state $s_t \in \gS$ and generates a stochastic joint observation $o_t = (o_t^1, \cdots, o_t^k) \sim O(\cdot | s_t)$, which is added to the action-observation trajectory $\tau_t = (o_0, a_0, \cdots, o_{t-1}, a_{t-1}, o_t)$. Each drone agent $j \in \gN$ then selects an action $a_t^j \in \gA^j$ using its policy $\pi^j(a^j | \tau_t^j)$. The environment transitions to state $s_{t+1}$ with probability $\gP(s_{t+1} | s_t, a_t)$, and all pursuers receive a common reward $r(s_t, a_t)$. Considering the discount factor $\gamma \in [0, 1]$, the discounted return is $R(\tau) = \sum_{t=0}^T \gamma^t r(s_t, a_t)$. The objective is to maximize the learner’s expected return with the sampled teammates, formally given by $J = \E_{\pi^{-1} \sim \pi^{\gU}} \E_{\tau \sim {\pi^1, \pi^{-1}, \pi^e}} R(\tau)$.
The zero-shot multi-drone pursuit problem extends conventional zero-shot coordination~\cite{otherplay,carroll2019utility,COLE} beyond two-player settings to multi-player scenarios. 
Besides, unlike traditional approaches that involve only teammates, this problem introduces the additional challenge of adversarial agents (evaders), requiring drones to coordinate with unseen teammates while simultaneously pursuing opponents.

\section{Hypergraphic Open-ended Learning Algorithm}

To address the zero-shot multi-drone pursuit problem, we introduce a novel approach named the hypergraphic open-ended learning algorithm (\algo). 
In Section~\ref{sec:Pref_H}, we first introduce the preference hypergraph and hyper-preference centrality to model cooperative relationships and assess the coordination ability of each agent within the hypergraph.
In Section~\ref{sec:HOLA}, we provide the details of \algo, as illustrated in Fig.~\ref{fig:model}.

\begin{figure}[t]
  \centering
    \includegraphics[width=\linewidth]{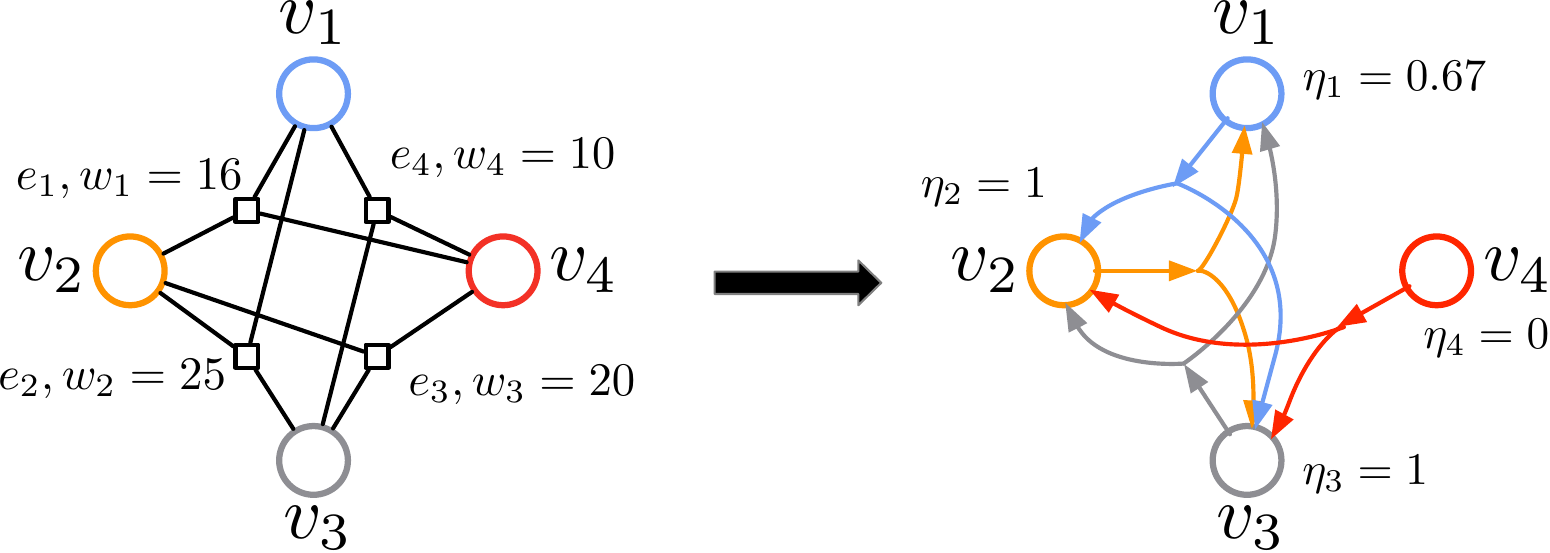} 
    \caption{Schematic diagrams of hypergraph representation of HyFog (left figure) and its preference hypergraph (right figure). The hyper-preference centrality is calculated using in-degree centrality.}
    \label{fig:pref_H}
\end{figure}

\subsection{Preference Hypergraph}

\label{sec:Pref_H}

We first propose the hypergraphic-form game to model the interactions in a population of agents. 

\begin{definition}[Hypergraphic-Form Game]
    The \textbf{Hy}pergraphic-\textbf{Fo}rm \textbf{G}ame (HyFoG) $\gG$ is defined by tuple $(\gV, \mathcal{E}, \vw, l)$. $\mathcal{V}$ is a finite set of vertices representing players, each indexed by $i$ and parameterized by the weights of a neural network. $\mathcal{E}$ is a set of hyperedges with fixed length $l$.
    $\vw$ is a weight vector consisting of the utility obtained for each hyperedge connected nodes co-playing.
\end{definition}

\begin{figure*}[ht]
    \centering
    % \vspace{-0.2cm}
    \includegraphics[width=\textwidth]{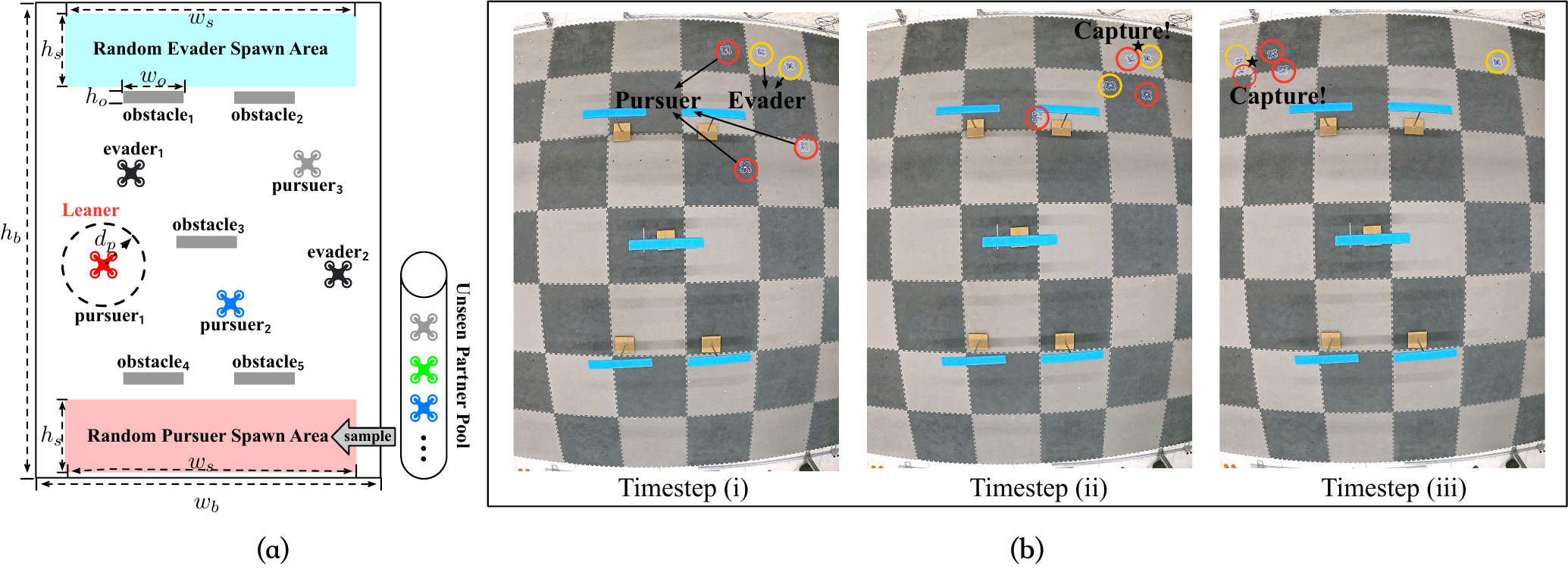}
        % \vspace{-0.1cm}
        \caption{
        (a) Schematic diagram of the cooperative drone pursuit environment with 3 pursuers and 2 evaders. (b) Snapshots of the real-world experiment from the top view. In timestep (ii) and timestep (iii), the pursuers successfully capture the two evaders, respectively.
        }
        \label{fig:system}
        % \vspace{-0.7cm}
\end{figure*}

The HyFog exhibits the following two properties. HyFog is \textit{l-uniform} as $|e|=l$ for all $e\in \mathcal{E}$. HyFog is \textit{connected} as every pair of nodes is connected. 
Although the HyFog model offers a detailed framework for agent interactions, directly extracting data relevant to cooperative capabilities within the game remains a challenge. 
To address this, we further introduce the  preference hypergraph $\mathcal{P}\mathcal{G}$.

\begin{definition}[Preference Hypergraph]
A preference hypergraph, denoted as $\mathcal{PG}$, is an unweighted directed hypergraph derived from a HyFog $\mathcal{G} = (\mathcal{V}, \mathcal{E}, \mathbf{w})$ and represented as $(\mathcal{V}, \widetilde{\mathcal{E}})$. In $\mathcal{PG}$, for each node $i \in \mathcal{V}$, there exists a unique outgoing hyperedge $\widetilde{e}_i$ such that $w(\widetilde{e}_i) = \max_{e \in \mathcal{E}_i} w(e)$, where $\mathcal{E}_i$ is the set of hyperedges that connect to $i$.
% This hyperedge connects node $i$ to $l-1$ other nodes in $\mathcal{V}$, forming a team that achieves the highest outcome for node $i$ among all possible coalitions within $\mathcal{V}$.
\end{definition}

In hyper-preference graph, a node that is the endpoint of multiple hyperedges typically indicates a higher cooperative ability.
We introduce the concept of \bt{hyper-preference centrality}, denoted by $\eta$, to quantify the cooperative ability of each node. For any node $i \in \mathcal{V}$, the hyper-preference centrality $\eta_i$ is defined as
\begin{equation}
    \eta_i = \frac{1}{|\gV|-1}d(i),
\end{equation}
where $d(i)$ is a centrality metric that quantifies the importance or influence of node $i$ within the network. 

Intuitively, each hyperedge in $\widetilde{\mathcal{E}}$ signifies a preference relationship, with the source node favoring the formation of a coalition with the end nodes. This preference arises because the source node achieves the highest outcomes when co-playing with the end nodes. Fig.~\ref{fig:pref_H} provides a schematic illustration of the hypergraph representation of HyFog (left) and its corresponding preference hypergraph (right). In this example, nodes $v_2$ and $v_3$ exhibit the highest cooperative capacity, each with a hyper-preference centrality value of 1.

\subsection{Hypergraphic Open-ended Learning Algorithm}

\label{sec:HOLA}
% Figure~\ref{fig:model} provides a detailed illustration of one generation $j > 0$ within the open-ended learning process. 
% In each generation $j$, the algorithm takes the learner and HyFoG generated by the previous generation as input. Then it trains a best-preferred agent and generates the corresponding HyFoG for the next generation. This iterative process ensures that the coordination capabilities of the agents are continuously improved over successive generations.
% \vspace{-1cm}
We then incorporate the hypergraphic-form game into open-ended learning framework and propose \algo, which continuously adjusts training objectives to enhance coordination capabilities among agents. \algo consists of two main phases: the pre-training phase and the open-ended learning phase.

\paragraph{Pre-training Phase.} To improve the diversity of policies in the hypergraph, we first pre-train a population of drone agents and then construct the initial HyFoG $\gG_0$. 
Motivated by maximum entropy RL and MEP~\cite{MEP}, we incorporate an additional maximum entropy goal into the training objective function. This encourages the development of a population of agents that can cooperate effectively while employing mutually distinct strategies.
The objective function is defined as follows:
\begin{equation}
J(\bar{\pi}) = \sum_t \mathbb{E}_{\left(s_t, a_t\right) \sim \bar{\pi}}\left[R\left(s_t, a_t\right) + \alpha \mathcal{H}\left(\bar{\pi}\left(\cdot \mid s_t\right)\right)\right],
\end{equation}
where $\bar{\pi}$ represents the mean policy of the population, and $\alpha \in [0,1]$ is a balancing constant. This formulation aims to maximize the expected return while simultaneously encouraging policy diversity through the entropy term:
\begin{equation}\mathcal{H}\left(\bar{\pi}\left(\cdot \mid s_t\right)\right)=-\sum_{a \in \mathcal{A}} \bar{\pi}^{(i)}\left(a_t \mid s_t\right) \log \bar{\pi}^{(i)}\left(a_t \mid s_t\right).
\end{equation}

\paragraph{Open-ended Learning Phase. } After pretraining the initial HyFoG, \algo begins continuous training by developing a best-preferred agent over the current HyFoG. 
\bt{Best-preferred agent} is defined by the hyper-preference centrality $\eta=1$. 
Intuitively, the best-preferred agent $j$ for agent $i$ ($i \neq j$) is the one with whom agent $i$ can achieve the highest reward compared to all other agents in the population. 
This reflects the agent’s best adaptive coordination ability within the group.
This process aims to continuously improve the cooperative ability among diverse agents.

The first module of \algo is the \bt{Grapher module}, which builds the latest HyFoG with the newly generated Learner agent. 
Grapher constructs each hyperedge between the newly generated Learner and the $l-1$ nodes sampled from the previous HyFoG. 
It then simulates interactions to obtain the average result as the weight of each hyperedge. Upon completion, this module provides the latest HyFoG and the corresponding hyper-preference graph.

The next module, \bt{Oracle module}, is the core component of the \algo algorithm. 
Its purpose is to train the best-preferred agent to the newly constructed HyFoG. Formally, given strategy $j$ and the current HyFoG $\gG_{j}$, the Oracle returns a new strategy $j+1$:
\begin{equation}
    j+1 = \br(j, \gG_{j}, \gJ_j), \textit{ with } \eta(j+1)=1,
    \label{eq:oracle}
\end{equation}
where $\gJ_j$ is the objective function.
The \bt{objective function} $J_j$ is defined as follows:
\begin{equation}
    J_j = \E_{\pi^{-1}_j \sim \phi(\gV_j)} \E_{\tau \sim \{\pi^1_j, \pi^{-1}_j, \pi^e\}} \left[ R(\tau) \right],
    \label{eq:obj}  
\end{equation}
where $\pi^{-1}_j$ represents the policies of other agents in the population sampled from the set of vertices $\gV_j$ according to the distribution $\phi$. 

The distribution $\phi$ is derived by the $\bm{\phi}$ \bt{Solver submodule}, which measures the cooperative ability of each node in the HyFoG. Agents with lower cooperative ability in the hypergraph are assigned a higher probability in the distribution $\phi$. Thus, the objective function specifically encourages improved cooperative performance with those fail-to-collaborate agents, enhancing overall coordination ability within the population.
Specifically, we adapt the inverse of the Myerson value to calculate $\mathcal{\phi}$ in HyFoG. The Myerson value $\phi^{-1}$ for any player $i\in \gV$ in HyFog is calculated as follows:
\begin{equation}
    \label{eq:mv}
    \phi^{-1}_i = \frac{1}{|\Pi(\gN)|}\sum_{\sigma \in \Pi(\gN)}[v(\gP_i^\sigma \cup \{i\}) - v(\gP_i^\sigma)],
\end{equation}
where $\Pi(\gN)$ is the set of all permutations defined on $\gN$. 
The derivation of Eq.~\ref{eq:mv} is provided in Appendix~\ref{app:mv}.

\begin{table}[t]
        \centering
        \caption{Parameters of experiment setting.}
        \begin{tabular}{c|c|c}
            \toprule
            \textbf{Param.} & \textbf{Values} & \textbf{Description}\\ \midrule
            $w_b$ & 3.6 m & Boundary Width\\ 
            $h_b$ & 5 m  & Boundary Height \\ 
            $w_s$ & 3.2 m & Spawn Area Width \\ 
            $h_s$ & 0.6 m & Spawn Area Height \\ 
            $w_o$ & 0.65 m & Obstacle Width \\ 
            $h_o$ & 0.1 m & Obstacle Height \\
            \midrule
            $d_c$ & 0.2 m & Capture Distance\\ 
            $d_p$ & 2 m & Perception Range\\ 
            $d_s$ & 0.1 m & Drone Safe Radius\\ 
            $v_P$ & 0.3 m/s & Velocity of Pursuers\\ 
            $v_E$ & 0.6 m/s & Velocity of Evaders \\
            \midrule
            $t_{max}$ & 100 s & Task Horizon \\
            fps & 10 & Frames Per Second\\ 
            \midrule
            $\mathcal{U}$ & $\{u_1,\dots,u_4\}$ & Unseen Partner Pool
            \\ 
            \bottomrule
        \end{tabular}
        \label{tab:arena_setting}
    \end{table}

The \bt{constraint} $\eta(j+1)=1$ in objective function ensures that the new strategy $j+1$ is the best-preferred agent, indicating that it has the highest preference centrality within the group, thus demonstrating a better cooperative ability. 
However, achieving the condition $\eta(j+1) = 1$ is not always feasible due to the strictness of this requirement. Thus, in the \algo framework, we modify the condition to require that the preference centrality ranking of $j+1$ be within the top $m$. The new \abr\ is defined as:
\begin{equation}
    j+1 = \abr(j, \gG_{j}, \gJ_j), \textit{ with } f(\eta(j+1)) > m,
\end{equation}
where $f(\cdot)$ is the ranking function. The strategy $j+1$ is then called \bt{\abp}.

\section{Experiment}

In this work, we conducted a series of experiments to verify the effectiveness of our proposed \algo method in coordinating with unseen partners. Section \ref{sec:exp_setting} introduces the experimental setup, metrics, baselines, and other relevant details. This is followed by an explanation of the unseen drone team configurations in Section \ref{sec:exp_unseen}. The results of the experiments and their analysis are presented in Section \ref{sec:exp_res}.

\subsection{Experiment Setting}
\label{sec:exp_setting}

\paragraph{Cooperative Drone Pursuit Environment.} The experiments are carried out in a cooperative drone pursuit environment that features 3 pursuers and 2 evaders, as shown in the left of Fig.~\ref{fig:system}.
All drones operate within a rectangular area with a boundary width ($w_b$) of 3.6 meters and a boundary height ($h_b$) of 5 meters. 
The task horizon ($t_{max}$) is the maximum duration of each episode, set to 100 seconds. The simulation runs at 10 frames per second (fps), ensuring smooth and continuous tracking of drone movements.

At the start of each episode, the three pursuers ($p_1$, $p_2$, $p_3$) and two evaders ($e_1$, $e_2$) are randomly spawned in their designated areas. The spawn area for each group measures 3.2 meters in width ($w_s$) and 0.6 meters in height ($h_s$). The sky-blue rectangle in Fig.~\ref{fig:system} indicates the evaders’ spawn area, while the red rectangle indicates the pursuers’ spawn area.
    To introduce additional complexity, the arena features five obstacles, each with a width ($w_o$) of 0.65 meters and a height ($h_o$) of 0.1 meters. These obstacles are strategically placed to influence the movement dynamics of both pursuers and evaders.
    Several critical parameters influence the drones’ interactions. The capture distance ($d_c$) is set to 0.2 meters, which is the threshold distance within which a pursuer is considered to have captured an evader. The perception range ($d_p$) is 2 meters, defining the radius within which a drone can detect others. Each drone also has a safe radius ($d_s$) of 0.1 meters to avoid collisions.
    The pursuers move at a velocity ($v_P$) of 0.3 meters per second, while the evaders move faster, at a velocity ($v_E$) of 0.6 meters per second. 
    This difference in speed necessitates strategic coordination among the pursuers to successfully capture the evaders.
    Finally, the unseen partner pool ($\mathcal{U}$) consists of a set of strategies denoted as $\{u_1,\dots,u_4\}$, which are used to test the zero-shot coordination capabilities of the pursuers when teamed with previously unseen partners.

\paragraph{Physical Environment.} To verify our proposed \algo algorithm beyond simulation, we deploy the learned policies of \algo with unseen drone teammates in the multi-quadrotor system Crazyflie. We use the OptiTrack motion capture system to measure the positions and orientations of both pursuers and evaders. Our multi-quadrotor communication framework is based on CrazySwarm~\cite{crazyswarm}.
Snapshots of real-world experiments are shown in right of Fig.~\ref{fig:system}.

\paragraph{Evader Policy.} The evaders are controlled by the escape policy proposed by \cite{Janosov2017Group} and \cite{ZhangDACOOP2023}. This policy defines multiple repulsive forces exerted by the pursuers and obstacles on the evaders. Additionally, wall-following rules are incorporated to help evaders maneuver along obstacle surfaces when positioned between pursuers and obstacles.

\paragraph{Baselines and Metrics.} We use the task success rate, the collision rate, and the mean episode length as metrics to evaluate performance in coordinating with unseen partners.
The unseen partners will be introduced in further detail in Section~\ref{sec:exp_unseen}.
An episode is deemed \bt{successful} if the pursuers capture both evaders, defined as reducing the distance between an evader and a pursuer to less than 0.2 meters. A \bt{collision}
is recorded if the distance between any two pursuers is less than 0.2 meters or if the distance between a drone and an obstacle is less than 0.1 meters. To evaluate the \algo algorithm, we compare it with self-play (SP)~\cite{schulman2017equivalence,carroll2019utility}, population-based training (PBT)~\cite{jaderberg2017population,carroll2019utility}, fictitious co-play (FCP)~\cite{FCP}, and maximum entropy population-based training (MEP)~\cite{MEP}. All methods, including \algo, are implemented using the PPO algorithm~\cite{PPO}. The action space is continuous, ranging from 0 to 1, and determines the drone's direction, while the drone's velocity remains fixed and predefined.
Further details of implementation are available in Appendix~\ref{app:implementation}.

\begin{table}[t]
\centering
\caption{
One-evader capture success rate (SR) and average episode length (AEL) performance of drone agents in unseen teammate pools: Heterogeneous Pool and Homogeneous Pool. 
The numbers (1) and (2) following PPO and D3QN-G represent models trained with different seeds. 
All results are averaged over 50 validation episodes.
}
\resizebox{\linewidth}{!}{
\begin{tabular}{ccccccc}
    \toprule
    \multirow{2}{*}{\textbf{Metrics}} & 
    \multicolumn{2}{c}{\textbf{Homogeneous}}& \multicolumn{4}{c}{\textbf{Heterogeneous }}  \\
    \cmidrule(lr){2-3} \cmidrule(lr){4-7}
     & \textbf{PPO (1)} & \textbf{PPO (2)} & \textbf{Greedy} & \textbf{VICSEK} & \textbf{D3QN-G (1)} & \textbf{D3QN-G (2)}  \\
    \midrule
    \textbf{SR} & 90.0\% & 72.0\% & 62.0\% & 98.0\% & 80.0\% & 78.0\%  \\
    \textbf{AEL} & 321.94 & 466.22 & 561.78 & 295.88 & 435.78 & 510.34  \\
    \bottomrule
\end{tabular}
}
% \vspace{-0.3cm}
\label{tab:unseen_pool}
\end{table}

\begin{figure*}[t]
    \centering
    \vspace{-0.5cm}\includegraphics[width=0.9\linewidth]{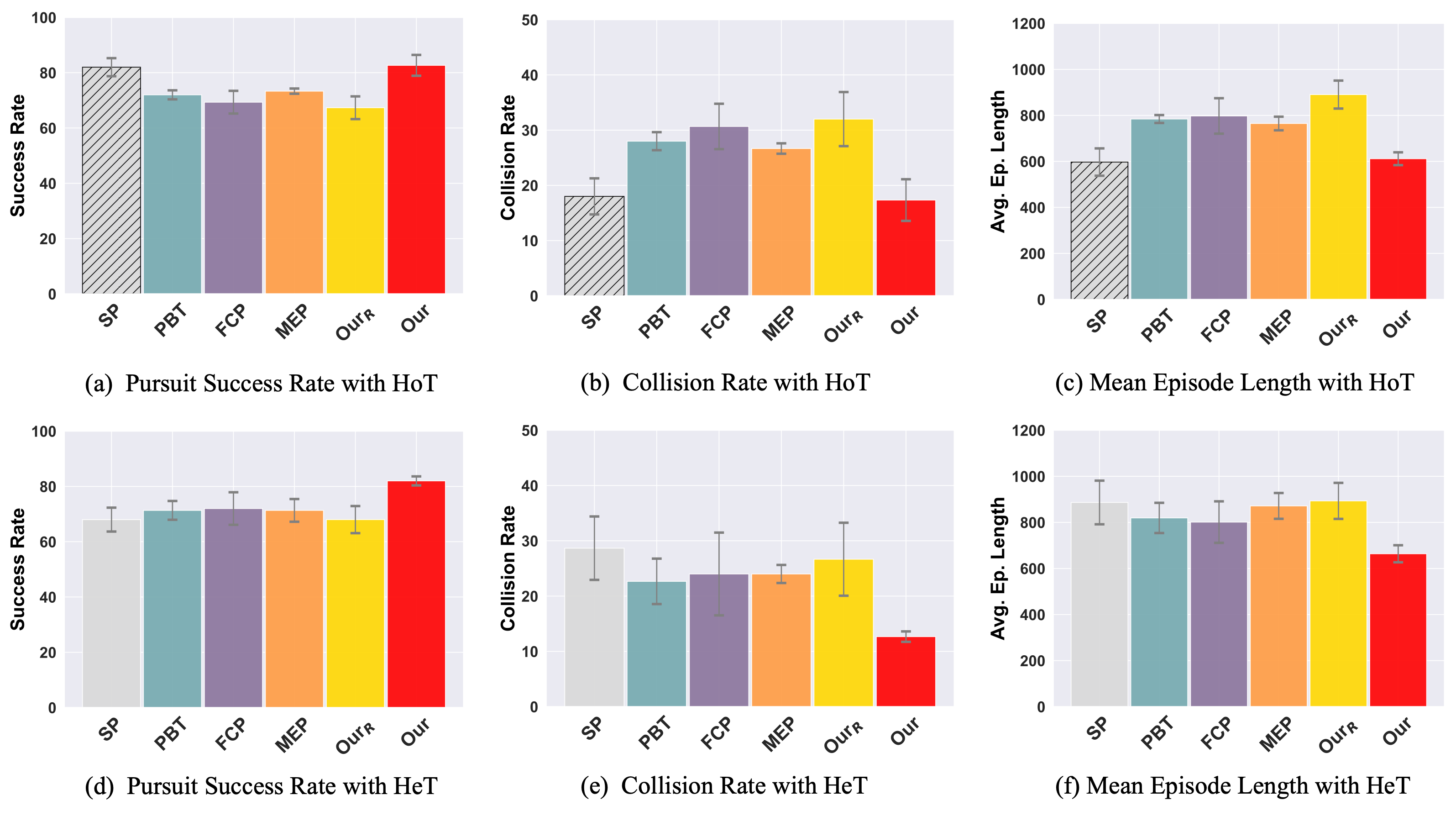}
        \caption{
    Comparison of Task Success Rate (first column, higher is better), Collision Rate (second column, lower is better), and Mean Episode Length (third column, lower is better) among four baseline methods, one ablation method \algoR, and our proposed \algo in the 3-Pursuer-2-Evader Scenario when playing with both Homogeneous Teammates (HoT) and Random Heterogeneous Teammates (HeT).
    \textbf{The first row} depicts the performances with two homogeneous teammates.
    \textit{The results for the SP method (slashed bar) are obtained from co-playing with the same algorithms and should be excluded from the comparison.}
    \textbf{The second row} shows the results obtained from co-playing with random teammates sampled from an unseen teammate pool. 
    The means and standard deviations, indicated by the error bars, are calculated over three different random seeds, with each seed undergoing 50 repeated runs.
    }
    %     \caption{
    % Comparison of Task Success Rate (first column, higher is better), Collision Rate (second column, lower is better), and Mean Episode Length (third column, lower is better) among four baseline methods, one ablation method \algoR, and our proposed \algo in the 3-Pursuer-2-Evader Scenario when playing with (random) Heterogeneous Teammates (HT).
    % \textbf{The first row} depicts the performances with two homogeneous teammates, both trained under self-play (SP) with different random seeds. \textit{The results for the SP method (slashed bar) are obtained from co-playing with the same algorithms.}
    % \textbf{The second row} shows the results from co-playing with random HT sampled from an unseen teammate pool. 
    % The means and standard deviations (error bars) are calculated over three different random seeds, with each seed undergoing 50 repeated runs.
    % }
    % \vspace{-5mm}
    \label{fig:main_res}
\end{figure*}

\subsection{Unseen Drone Teammate Pools}

\label{sec:exp_unseen}

To evaluate the cooperative capabilities of our proposed \algo algorithm with previously unencountered drones, we establish two separate unseen drone teammate pools - homogeneous pool and heterogeneous pool.
The agents in the homogeneous pool are trained using a self-play PPO algorithm~\cite{PPO,carroll2019utility} with two different seeds. Table~\ref{tab:unseen_pool} presents the mean success rate for capturing an evader (SR) and the average episode length (AEL) of the two PPO agents when co-playing with themselves. The PPO (1) agent is an expert pursuer, achieving a 90\% SR and a shorter AEL. In contrast, the PPO (2) agent is a medium-level pursuer with a 72\% SR.

To verify cooperative ability with different algorithms, we establish a heterogeneous unseen drone agent pool consisting of four models: a Greedy agent, a VICSEK agent~\cite{Janosov2017Group}, and two D3QN-G agents. 
The D3QN-G agent is an ensemble algorithm that combines the Double Deep Q-Network (D3QN)~\cite{wang2016dueling} with the Greedy strategy. This approach is based on our experimental findings, which showed that while D3QN alone struggles with the 3-Pursuer-2-Evader task, it can effectively handle it when combined with the Greedy strategy.
More details of four models are provided in Appendix~\ref{app:unseen}. As shown in Table~\ref{tab:unseen_pool}, the ability of these agents ranges from the low-level Greedy strategy with a 62\% SR, to the medium-level D3QN-G models with approximately 80\% SR, and the expert-level VICSEK strategy with a 98\% SR. At each evaluation, we randomly sample two teammates from the pool to test zero-shot coordination ability. This approach allows us to comprehensively evaluate cooperative pursuit performance with teammates of varying skill levels.

\subsection{Experiment Results}

\label{sec:exp_res}
Fig.~\ref{fig:main_res} provides a comprehensive comparison of the coordination abilities of the proposed \algo algorithm, four baselines (SP, PBT, FCP, and MEP), and an ablation model \algoR, which removes the $\phi$ Solver module and uses the inverse mean reward as $\phi$. 
The results of the experiment verify the effectiveness of $\phi$ Solve module and demonstrate the our proposed \algo algorithm achieves better performance when cooperating with unseen teammates in both homogeneous and heterogeneous settings compared to the baseline methods.
Additionally, the real-world experiment further validates the feasibility of \algo in physical systems, as shown in Fig.~\ref{fig:main_res} and in the videos on our website.

\paragraph{Ablation Study: the Effectiveness of $\phi$ Solver Module.} To evaluate the effectiveness of the $\phi$ Solver module within the \algo algorithm, we conduct an ablation study by removing the $\phi$ Solver, named \algoR. Instead, we use the inverse of the mean reward of each node achieved with other nodes as the distribution $\phi$. The results, depicted in Fig.~\ref{fig:main_res}, demonstrate a notable difference in performance between the \algo (red bar) and the ablation model \algoR (gold bar). 
When the $\phi$ Solver module is removed, we observe a decrease in the Pursuit Success Rate, as well as an increase in the Collision Rate and the Mean Episode Length, for both homogeneous and random heterogeneous teammate configurations.

\paragraph{Coordination with Homogeneous Teammates.}
The first row of Fig.~\ref{fig:main_res} shows the performance across three metrics when cooperating with unseen homogeneous teammates. 
The results demonstrate the superior performance of \algo compared to other baselines when coordinating with unseen homogeneous teammates.
The results for the SP method are specifically marked with slashes, indicating that these results are obtained by co-playing with the same algorithm, unlike other methods that co-play with entirely different algorithms.
As a result, SP achieves the best overall performance compared to other baselines and \algo. For instance, the mean episode length of SP is 597.09, which is lower than the second-lowest method, \algo, at 611.58. Under the same evaluation criteria for coordination with unseen partners, \algo outperforms the other baselines. Compared to the second-best method, MEP, \algo shows about a 10\% improvement in the pursuit success rate. Additionally, the mean episode length of \algo is 611.58, significantly shorter than second-best model MEP’s 764.75.

\paragraph{Coordination with Random Heterogeneous Teammates.}
In addition to evaluating with homogeneous unseen teammates, we assess the cooperative ability with random heterogeneous teammates. As described in Section~\ref{sec:exp_unseen} and Appendix~\ref{app:unseen}, the heterogeneous unseen pool consists of four different drone agents whose cooperative abilities range from medium to expert levels. At the beginning of each evaluation, we randomly sample two teammates from this pool. This process is repeated 50 times for each seed. The bottom figures in Fig.~\ref{fig:main_res} compare the performance of \algo with other baselines across three metrics.
Compared to the baseline methods, \algo achieves the best performance in all three metrics. The mean episode lengths for all four baselines are greater than 800, with the shortest being the FCP method at 801.19. In contrast, the mean episode length for \algo is significantly shorter at 663.72.
Additionally, \algo outperforms the baselines in terms of success and collision rates, with a success rate of 82\% and a collision rate of 13.67\%. This is compared to the second-best method, FCP, which has a success rate of 72\%, and PBT, which has a collision rate of 22.67\%.
When coordinating with more complex and diverse unseen teammates, \algo achieves more significant improvements than in the homogeneous teammates setting.
\section{Conclusion}
In this paper, we propose a hypergraphic open-ended learning algorithm (\algo) to address the zero-shot cooperative multi-drone pursuit problem, enabling coordination with unseen drone partners. To the best of our knowledge, this is the first work to formulate the cooperative multi-drone pursuit task as zero-shot multi-agent coordination problem within the Dec-POMDP framework. This formulation extends ZSC research from two-player video games to real-world multi-drone cooperative pursuit scenarios.
We introduce a novel hypergraphic open-ended learning algorithm that continuously enhances the learner’s cooperative ability with multiple teammates. To empirically verify the effectiveness of \algo in coordinating with unseen teammates, we construct two unseen drone teammate pools for evaluation, comprising both homogeneous and heterogeneous teammates. Experimental results in both simulation and a real-world cooperative Crazyflie pursuit environment demonstrate that \algo can better coordinate with unseen teammates compared to baseline methods.

\textbf{Limitations and Future Work:} While our study demonstrates the effectiveness of \algo in achieving zero-shot coordination with unseen drone partners, several limitations warrant further investigation. Firstly, although collision avoidance is considered in the reward function, some pursuers exhibit aggressive and dangerous behaviors to capture evaders. Future work should focus on incorporating more sophisticated safety mechanisms to ensure robust and reliable performance in real-world applications.
While the deployment of \algo policies in Crazyflie 2.1 drones validates its feasibility in physical systems, scalability remains a challenge. Future research could explore optimizing \algo for larger swarms and more complex tasks.
Finally, integrating advanced sensors and communication protocols could enhance the coordination and efficiency of drone swarms, addressing the current variability in hardware capabilities and further bridging the gap between simulation and real-world applications.

%===============================================================================

% The acknowledgments are automatically included only in the final and preprint versions of the paper.
% \acknowledgments{If a paper is accepted, the final camera-ready version will (and probably should) include acknowledgments. All acknowledgments go at the end of the paper, including thanks to reviewers who gave useful comments, to colleagues who contributed to the ideas, and to funding agencies and corporate sponsors that provided financial support.}

%===============================================================================

% no \bibliographystyle is required, since the corl style is automatically used.
\bibliographystyle{ieeetr}
\bibliography{example}  % .bib

%appendix
\section{Appendix}
This appendix provides additional information and supporting materials that complement the main content of the paper.

\label{app:system}
    \begin{table*}[t]
        \centering
        \caption{Implementation hyperparameters of \algo.}
        \begin{tabular}{@{}cc|cc@{}}
            \toprule
            \textbf{Parameters} & \textbf{Values} & \textbf{Parameters} & \textbf{Values}\\ \midrule
            Batch size & 1024 & Minibatch size & 256\\ 
            Lambda (\(\lambda\)) & 0.99  & Generalized advantage estimation lambda (\(\lambda_{gae}\)) & 0.95\\ 
            Learning rate & 3e-4 & Value loss coefficient($c_1$) & 1 \\ 
            Entropy coefficien t(\(\epsilon_{clip}\)) & 0.01 & PPO epoch & 20 \\ 
            Total environment step & 1e6 & Max hypergraph size & 10 \\ 
            \bottomrule
        \end{tabular}
        \label{tab:algo_params}
    \end{table*}
    
\subsection{Derivation of Eq.~\ref{eq:mv}} 
\label{app:mv}

In the module of $\phi$ Solver,  we adapt the inverse of the Myerson value to calculate $\mathcal{\phi}$ in HyFoG. The Myerson value $\phi^{-1}$ for any player $i\in \gV$ in HyFog is calculated as follows:
\begin{subequations}
\begin{align}
    \phi^{-1}_i &= SV_i(\gN, v^\gE) \\
    &= \frac{1}{|\Pi(\gN)|}\sum_{\sigma \in \Pi(\gN)}[v^\gE(\gP_i^\sigma \cup \{i\}) - v^\gE(\gP_i^\sigma)] \\
    &= \frac{1}{|\Pi(\gN)|}\sum_{\sigma \in \Pi(\gN)}\left[\sum_{T \in \gP_i^\sigma \cup \{i\} \setminus \gE} v(T) - \sum_{T \in \gP_i^\sigma \setminus \gE} v(T)\right] \label{eq:mv1} \\
    &= \frac{1}{|\Pi(\gN)|}\sum_{\sigma \in \Pi(\gN)}[v(\gP_i^\sigma \cup \{i\}) - v(\gP_i^\sigma)] \label{eq:mv2}
\end{align}
\end{subequations}

The transition from Eq.~\ref{eq:mv1} to Eq.~\ref{eq:mv2} occurs because HyFoG is connected, ensuring that any subset $S \subseteq \gN$ is also connected. In this context, the components of the hypergraph $(\gP_i^\sigma \cup \{i\}, \gE)$ and $(\gP_i^\sigma, \gE)$ remain unchanged. For any coalition $S \subseteq \gN$, if $|S| < r$, the value of the coalition $v(S) = 0$. Otherwise, when $|S| \geq r$, $v(S) = \sum_{T \in \triangle(S)} \vw(T)$, where $\triangle(S)$ denotes the subset of $S$ with a fixed size of $r$.

\subsection{Implementation Details of \algo}
\label{app:implementation}

In this section, we outline the hyperparameters used in the implementation of \algo algorithm. The selection of these hyperparameters is crucial for the effective training and performance of the algorithm. Table \ref{tab:algo_params} summarizes the specific values for each parameter used in our experiments.

\subsection{Heterogeneous Unseen Drone Teammate Pool}
\label{app:unseen}

In this section, we will further introduce details of the heterogeneous unseen drone teammate pool, consisting of four models: a Greedy agent, a VICSEK agent, and two
D3QN-G agents. The details of the four agents are as follows:
\begin{itemize}
    \item \textbf{Greedy Agent.} The Greedy agent pursues the target independently, continually adjusting its movement to align with the target’s position. Its state information includes its own position and orientation, distances and angles to teammates and evaders, and proximity to obstacles or walls. If obstacles or other pursuers are detected within its evasion range, the agent adjusts its direction to avoid them.
    \item \textbf{VICSEK Agent.} Inspired by research on group chasing tactics~\cite{Janosov2017Group}, this strategy involves continuously computing and updating the velocity vector directed towards the evader based on the agent’s current environmental state to optimize the tracking path. When the agent detects potential obstacles or other chasers nearby, it automatically evades them by applying repulsive forces with varying magnitudes and coefficients. Although the final velocity vector includes both magnitude and orientation, only the orientation is implemented in this experiment.
    \item \textbf{D3QN-G Agent.} The D3QN-G agent is an ensemble algorithm that combines the Double Deep Q-Network (D3QN)~\cite{wang2016dueling} with the Greedy strategy. Initially, the D3QN-G agent employs the D3QN method to pursue one of the evaders. Once the first evader is captured, it switches to the Greedy strategy to capture the second evader. This approach is based on our experimental findings, which showed that while D3QN alone struggles with the 3-Pursuer-2-Evader task, it can effectively handle it when combined with the Greedy strategy. The action space consists of 24 artificial potential field with attention (APF-A) parameter pairs ($\lambda$, $\eta$), formed by the Cartesian product of 8 $\lambda$ candidate parameters and 3 $\eta$ candidate parameters, following the setting in \cite{ZhangDACOOP2023}. The parameter $\eta$ is used to calculate the repulsive force, while $\lambda$ is used to calculate the inter-robot force. The state dimension in the training environment is 9. In addition to the information mentioned in the Greedy strategy, it also includes a bit indicating whether the current agent is active. If a teammate captures the target, the teammate transitions from an active state to an inactive state, thereby ceasing movement.
\end{itemize}

\end{document}